\documentclass[10pt,twocolumn,letterpaper]{article}

\usepackage[pagenumbers]{cvpr} 
\usepackage{multirow}
\usepackage{multicol}
\usepackage{array}

\newcommand{\TODO}[1]{\textbf{\color{red}[TODO: #1]}}
\renewcommand{\TODO}[1]{}

\definecolor{cvprblue}{rgb}{0.21,0.49,0.74}
\usepackage[pagebackref,breaklinks,colorlinks,allcolors=cvprblue]{hyperref}

\def\confName{CVPR}
\def\confYear{2026}

\title{SparseSplat: Towards Applicable Feed-Forward 3D Gaussian Splatting with~Pixel-Unaligned Prediction}

\author{
Zicheng Zhang$^{1}$, Xiangting Meng$^{2}$, Ke Wu$^{1}$, Wenchao Ding$^{1*}$\\
$^{1}$Fudan University \quad $^{2}$ShanghaiTech University\\
{\small Email: zichengzhang23@m.fudan.edu.cn \quad $^{*}$Corresponding author}\\
{\small Project page: \url{https://victkk.github.io/SparseSplat-page/}}
}
\begin{document}

\twocolumn[{
\renewcommand\twocolumn[1][]{#1}
\maketitle

\begin{center}
    \centering
    \vspace{-25pt}
    \includegraphics[width=0.95\linewidth]{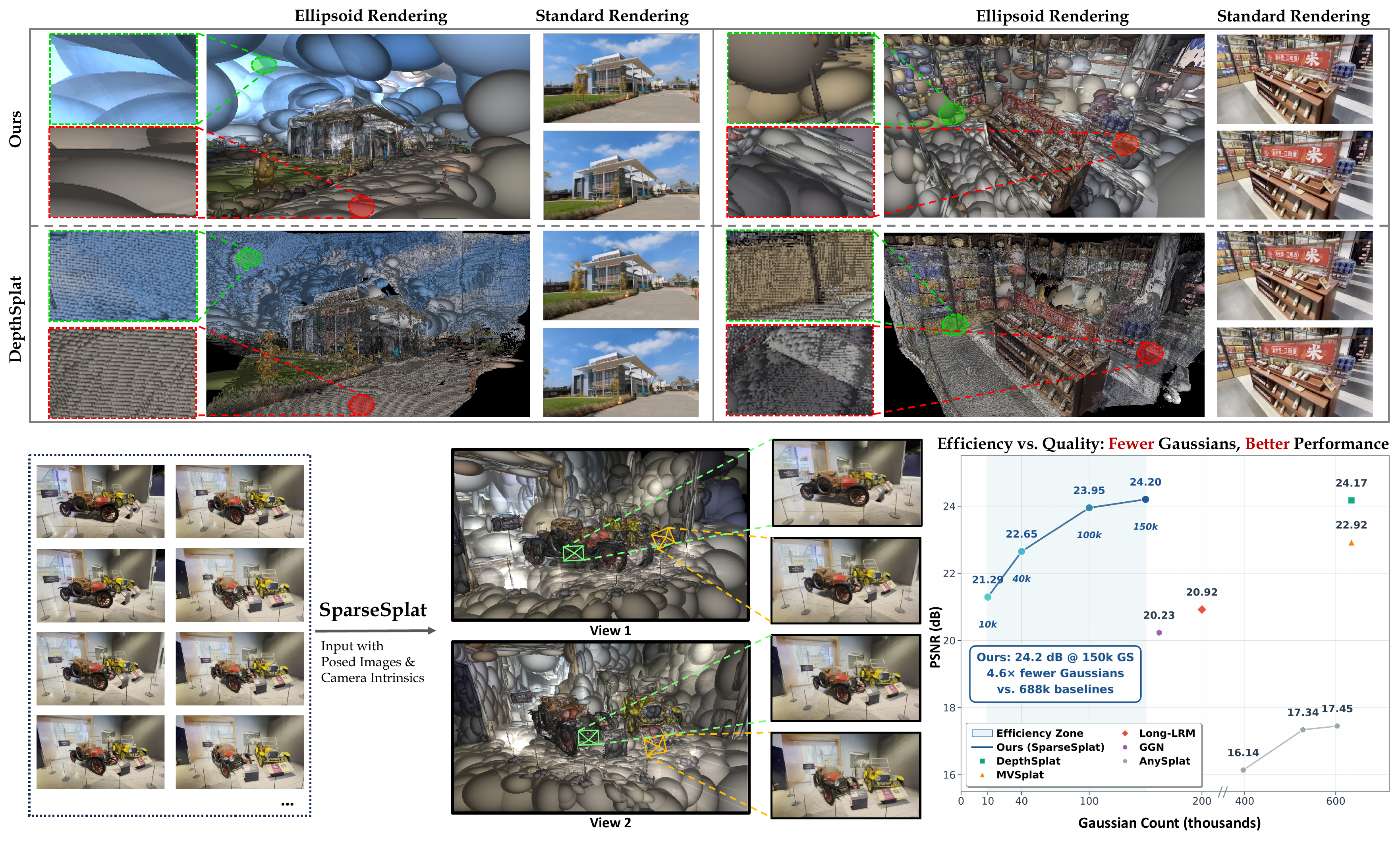}
    \captionof{figure}{SparseSplat achieves state-of-the-art rendering quality on DL3DV~\cite{dl3dv} using significantly fewer Gaussians than the previous SOTA, depthSplat~\cite{depthsplat} (150k vs. 688k). Our model also generates competitive results in sparse settings (e.g. 10k). As illustrated by ellipsoid renderings \cite{sibr2020}, SparseSplat adaptively allocates Gaussian density based on scene content. This contrasts with previous methods that adopt a pixel-aligned strategy, which produces spatially uniform and highly redundant Gaussian primitives even in textureless regions.}
    \label{fig:teaser}
\end{center}
}]
\begin{abstract}
Recent progress in feed-forward 3D Gaussian Splatting (3DGS) has notably improved rendering quality. However, the spatially uniform and highly redundant 3DGS map generated by previous feed-forward 3DGS methods limits integration into downstream reconstruction tasks. We propose SparseSplat, the first feed-forward 3DGS model that adaptively adjusts Gaussian density according to scene structure and information richness of local regions, yielding highly compact 3DGS maps. To achieve this, we propose entropy-based probabilistic sampling, generating large, sparse Gaussians in textureless areas and assigning small, dense Gaussians to regions with rich information. Additionally, we design a specialized point cloud network that efficiently encodes local context and decodes it into 3DGS attributes, addressing the receptive field mismatch between the general 3DGS optimization pipeline and feed-forward models. Extensive experimental results demonstrate that SparseSplat can achieve state-of-the-art rendering quality with only 22\% of the Gaussians and maintain reasonable rendering quality with only 1.5\% of the Gaussians.
\end{abstract}    
\section{Introduction}
\label{sec:introduction}
3D Gaussian Splatting (3DGS) \cite{3dgs} has recently emerged as a transformative technology for 3D scene representation and novel view synthesis. By representing scenes with millions of 3D Gaussian primitives, it achieves unprecedented rendering quality and real-time rendering speeds, surpassing prior methods like NeRF~\cite{nerf} and its accelerated variants \cite{yu2022plenoxels,mueller2022instant}. To accelerate its time-consuming optimization process, an up-and-coming research area is feed-forward 3DGS~\cite{pixelsplat,mvsplat,depthsplat,anysplat,longLRM}: training a model to generate a complete 3DGS scene representation in a single forward pass from a set of RGB images and camera poses. This "instant" reconstruction capability makes it an ideal choice for downstream tasks such as SLAM \cite{3dgsslam_survey,MonoGS,splatam,VingsMONO}, AR/VR \cite{ARVR_survey}, and robotics \cite{robot_survey}.

However, while recent works~\cite{mvsplat, depthsplat} have pushed rendering quality to new heights, the integration of feed-forward 3DGS into real-world downstream applications like SLAM remains limited. A core obstacle is the \emph{overly dense and redundant 3DGS maps} generated by current methods. This leads to significant memory and computation overhead, which is unacceptable for resource-constrained edge platforms such as cars, drones, or AR glasses.

The redundancy stems from the "pixel-aligned" nature. Early feed-forward methods typically predict one Gaussian primitive for each pixel in the input views, resulting in a uniformly distributed 3DGS map. This design is in sharp contrast to the adaptive representation generated by optimization-based 3DGS, which naturally uses sparse, large primitives in textureless areas and dense, small primitives in texture-rich regions. To mitigate this, the research community has made some initial explorations. AnySplat and VolSplat~\cite{anysplat, volsplat} explore voxelization, predicting one Gaussian for each 3D voxel instead of a 2D pixel. Others, like GGN and Long-LRM~\cite{GGN, longLRM}, apply post-processing to merge and prune primitives after the pixel-aligned prediction. However, these approaches fail to address the root cause. Voxel-based methods still adhere to a rigid grid, while post-processing merely cleans up redundancy rather than preventing it. Consequently, none of these methods fully unlock the adaptive efficiency inherent to 3DGS.

In this paper, we argue that this failure is rooted in two fundamental design flaws. First, a structural distribution mismatch. The methods above all enforce a rigid, uniform-by-nature structure (be it pixel or voxel), which is fundamentally incompatible with the content-aware, non-uniform distribution 3DGS is optimized for. Second, a receptive field mismatch. We observe that 3DGS optimization is an inherently local process (\cref{sec:prediction})—a Gaussian's attributes are determined by its immediate neighborhood in 2D and 3D space. However, existing methods~\cite{mvsplat,depthsplat} employ a contradictory design: they use backbones with vast receptive fields to extract global context, yet their prediction heads regress Gaussian attributes from a \textbf{single pixel's} feature. This mismatch between "global perception and then local prediction" deviates from the locality of the prediction task.

To address these two fundamental mismatches, we introduce SparseSplat, a new feed-forward 3DGS framework designed from the ground up to generate sparse, efficient, and high-fidelity scene representations. To resolve the distribution mismatch, we abandon the "pixel-aligned" paradigm. We introduce an Adaptive Primitive Sampling stage (\cref{sec:sampling}) that first quantifies local "information richness" using Shannon entropy. This information map is then used to perform probabilistic sampling, generating sparse and non-uniformly distributed 3D anchor points, thereby mimicking the adaptive distribution of optimized 3DGS. Critically, this sampling process is governed by a temperature parameter ($\tau$), which provides an intuitive and direct control to trade off 3DGS map size (i.e., memory) for rendering quality, adapting our method for diverse downstream tasks. To resolve the receptive field mismatch, we propose a 3D-Local Attribute Predictor (\cref{sec:prediction}). Instead of regressing from a single-pixel feature, our method operates directly in a 3D space. For each sparse anchor point, we query its K-Nearest-Neighbors (KNN) and feed this local neighborhood information into a lightweight prediction head to regress the final Gaussian attributes. This design better aligns the predictor's receptive field with the local optimization nature of 3DGS. Our contributions can be summarized as follows.
\begin{itemize}
    \item We are the first to identify and analyze two fundamental design flaws in existing feed-forward 3DGS: the \textbf{distribution mismatch} (rigid structure vs. content-aware distribution) and the \textbf{receptive field mismatch} (global context vs. local optimization).
    \item We propose \textbf{Adaptive Primitive Sampling}, a novel entropy-based strategy that discards the "pixel-aligned" paradigm to generate a sparse, content-aware set of 3DGS, addressing the root cause of map redundancy.
    \item We introduce a lightweight \textbf{3D-Local Attribute Predictor} that leverages 3D K-Nearest-Neighbors, aligning the network's design with the local nature of 3DGS optimization.
\end{itemize}
\section{Related Work}
\label{sec:related_work}
\subsection{Point-based Networks}
Early point cloud networks typically construct their encoders based on PointNet~\cite{qi2017pointnet} and its variants~\cite{qi2017pointnet++, qian2022pointnext, yang2018foldingnet}. By applying MLPs independently to each point and combining them with symmetric aggregation functions, they learn permutation-invariant global representations directly from unordered point sets. To better capture local geometric and topological structures, later work introduced graph and point convolution operators. For instance, DGCNN~\cite{dgcnn} dynamically builds a KNN graph in feature space and uses EdgeConv to model neighborhood structure, while KPConv~\cite{thomas2019kpconv} and PointCNN~\cite{li2018pointcnn} either explicitly place kernel points in 3D space or learn a transform to map irregular point sets into a latent space suitable for convolution. However, the tight coupling between encoder and decoder limits flexibility and generalization capacity~\cite{xie2021segformer}. With the success of self-attention in other domains \cite{vaswani2017attention}, PCT~\cite{guo2021pct} and Point Transformer~\cite{zhao2021point} perform self-attention within local neighborhoods and incorporate appropriate positional encodings, enabling unified modeling of local geometry and long-range dependencies without relying on regular grids.

\subsection{Feed-Forward 3DGS}
\label{subsec:related_work}
Feed-forward 3D Gaussian Splatting (3DGS) models \cite{pixelsplat,mvsplat,depthsplat,mvsplat360,longLRM,splatformer} aim to achieve instant 3D reconstruction. PixelSplat \cite{pixelsplat}, MVSplat \cite{mvsplat}, and DepthSplat \cite{depthsplat} adopt a pixel-aligned paradigm, generating significant redundancy and memory overhead as they allocate primitives uniformly even in textureless regions. Subsequent work has aimed to improve efficiency. Gaussian Graph Network \cite{GGN} proposed a post-pruning mechanism. AnySplat \cite{anysplat} and VolSplat \cite{volsplat} shifted from 2D pixel to 3D voxel, predicting a Gaussian primitive for each 3D grid. However, these methods still rely on a uniform, scene-agnostic structured grid (either pixel or voxel). This contrasts with the classic 3DGS \cite{3dgs} optimization pipeline, which adaptively allocates primitive density based on scene information richness. We propose a different approach. SparseSplat avoids structured grids by introducing information richness aware probabilistic sampling. The sampling process is guided by local information entropy, resulting in a "sparse-by-design" representation that mimics the adaptive nature of classic 3DGS in a feed-forward manner.
\begin{figure*}[t!]
  \centering
    \includegraphics[width=\linewidth]{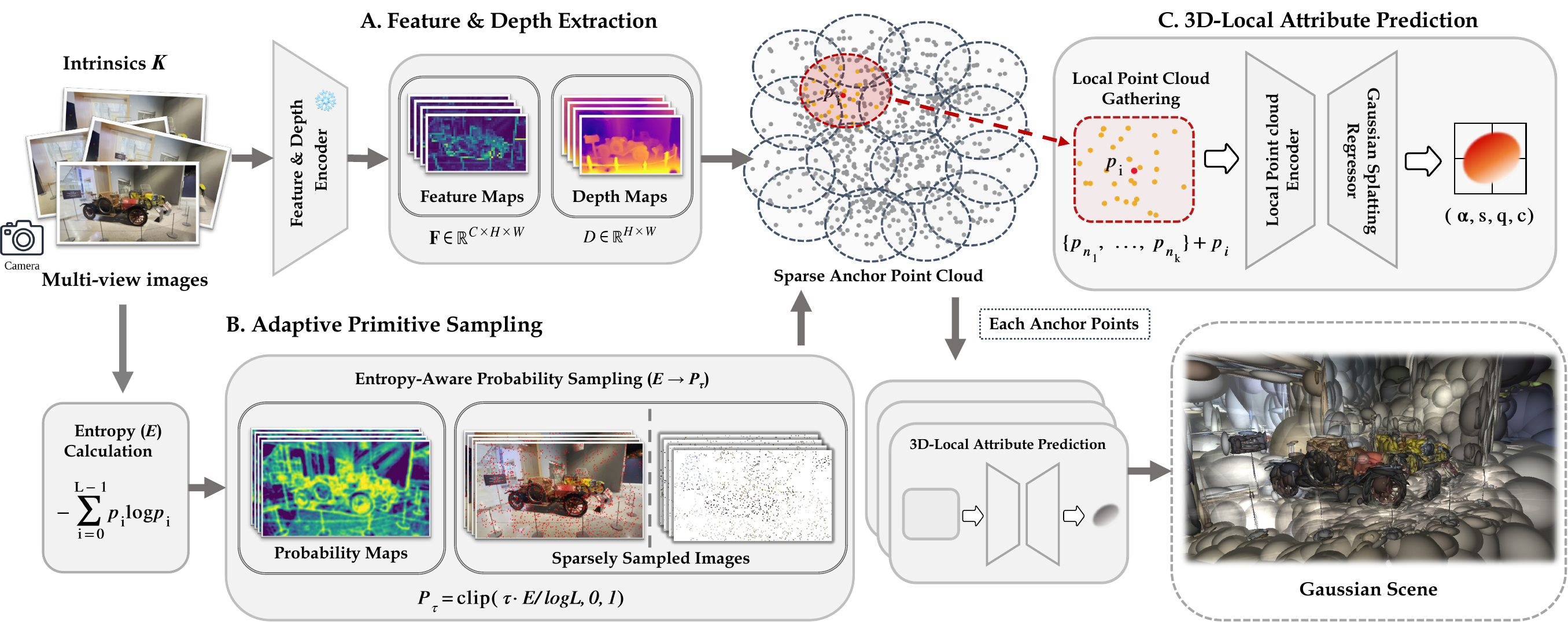}
    \vspace{-15pt}
    \captionof{figure}{\textbf{Overall Pipeline of SparseSplat.} Our method begins with using a frozen backbone \cite{depthsplat} to generate feature maps and depth maps from multi-view posed images. Next, in the Adaptive Primitive Sampling stage, the entropy maps are calculated and transformed into probability maps to perform sampling, resulting in sparse 2D pixels. These pixels are then back-projected into 3D space using the predicted depth to form 3D Sparse Anchor Points. Finally, for each anchor point, we gather its local point cloud neighborhood via KNN. This local neighborhood is fed into a lightweight prediction head to predict its complete Gaussian attributes ($\alpha, s, q, c$). All predicted Gaussian primitives are then merged to generate the final scene.}
    \label{fig:pipeline}
\end{figure*}

\section{Method}
\label{sec:method}

Our method, SparseSplat, is designed to generate a sparse and efficient 3D Gaussian scene representation from a set of posed RGB images in a single forward pass. Our core design adheres to the two key principles identified in the introduction: 1) to address the distribution mismatch, we adaptively allocate Gaussian primitives based on local information richness; and 2) to resolve the receptive field mismatch, we employ a local receptive field that matches the inherent nature of 3DGS optimization.

As illustrated in \cref{fig:pipeline}, our pipeline is divided into three main stages. First, in the \textbf{Feature and Depth Extraction} stage (\cref{sec:backbone}), we use a backbone network to extract 2D feature maps and depth maps from the input images, providing the foundation for subsequent geometry and appearance modeling. Second, in the \textbf{Adaptive Primitive Sampling} stage (\cref{sec:sampling}), we discard the "one-primitive-per-pixel" paradigm, introducing a novel, entropy-based probabilistic sampling strategy to generate sparse 3D anchor points only in information-rich regions. Finally, in the \textbf{3D-Local Attribute Prediction} stage (\cref{sec:prediction}), for each 3D anchor point, we query its K-nearest neighbors in 3D space and use a lightweight predictor to regress its complete Gaussian attributes based \emph{only} on this local neighborhood.

\subsection{Feature and Depth Extraction Backbone}
\label{sec:backbone}

Feed-forward 3DGS methods can typically be separated into two components, a depth prediction network and a Gaussian attribute predictor. The contribution of SparseSplat is not a new backbone architecture, but rather a new methodology for generating sparse Gaussian primitives.

Therefore, we adopt the \textbf{pre-trained} backbone network from~\cite{depthsplat} for multiview depth estimation~\cite{wang2024learningbasedmultiviewstereosurvey,yao2018mvsnet,yao2019recurrent,gu2020cascadecostvolumehighresolution,izquierdo2025mvsanywherezeroshotmultiviewstereo}. \textbf{During our training, the weights of this backbone are frozen}, and it is used solely to provide depth maps and 2D feature maps. For each input view $I$, this network generates a 2D feature map $\mathbf{F} \in \mathbb{R}^{C \times H \times W}$ and a depth map $D \in \mathbb{R}^{H \times W}$. Our method is backbone-agnostic, enabling it to directly benefit from future advancements in the field of multi-view depth estimation. Our novelty lies in how $\mathbf{F}$ and $D$ are utilized to generate a sparse scene representation.

\subsection{Adaptive Primitive Sampling}
\label{sec:sampling}

As established in the introduction, the "pixel-aligned" paradigm employed by current methods is the root cause of redundancy. It disregards the efficient distribution naturally achieved by classic 3DGS optimization~\cite{3dgs}: using sparse, large Gaussians in low-texture areas (like skies or walls) and dense, small Gaussians in texture-rich regions.

To address this, we propose an adaptive sampling mechanism based on local information entropy. The motivation is that the density of Gaussian primitives should be proportional to the local information complexity of the scene. The use of entropy to quantify image information, textural complexity, and irregularity is a well-established principle in image processing \cite{haralick1979statistical,rigau2003entropy}.

\paragraph{Information Richness Estimation.}We first quantify the local information richness at each pixel location. Specifically, for each input image $I$, we compute the Shannon entropy \cite{shannon1948mathematical} for each pixel within an $N \times N$ local window on its grayscale version, generating an information density map $E \in \mathbb{R}^{H \times W}$. High values in $E$ correspond to high-frequency, textured regions, while low values correspond to flat, low-frequency regions. The local entropy $E(u, v)$ is calculated as:
\begin{equation}
E(u, v) = - \sum_{i=0}^{L-1} p_i \log p_i\label{eq:entropy}
\end{equation}
where $L$ is the number of gray levels (e.g., 256), and $p_i$ is the normalized histogram count (probability) of gray level $i$ occurring within the window $\mathcal{W}_{u,v}$.

\paragraph{Probabilistic Sampling and 3D Back-projection.} The raw entropy map $E$ from \cref{eq:entropy} provides a measure of complexity. To convert them into a per-pixel sampling probability, we first normalize the map by its theoretical maximum (e.g., $\log256 = 8$) and then multiply by an adjustment coefficient $\tau$ controlling the number of points to be sampled. 

This process yields a probability map $P \in \mathbb{R}^{H \times W}$, where each $P(u, v)$ represents the independent probability of sampling the pixel at location $(u, v)$:
\begin{equation}
P_{\tau}(u, v) = \text{clip} \left( \tau \cdot \frac{E(u, v)}{\log L}, \ 0, \ 1 \right)
\label{eq:prob_map}
\end{equation}
where $\text{clip}(\cdot, 0, 1)$ ensures the value remains a valid probability. The coefficient $\tau$ serves as a crucial and intuitive control knob for sparsity. A higher $\tau$ increases the sampling probability across the board, leading to a denser point cloud, while a lower $\tau$ results in a sparser representation.

We then perform random sampling based on this probability map $P_{\tau}$. We iterate through every pixel $(u, v)$ in the $H \times W$ grid and generate a random number $r \sim U(0, 1)$. If $r < P(u, v)$, the pixel is selected and added to our sparse set $\mathcal{S}$.

Finally, for the pixels in the resulting sparse set $\mathcal{S}$, we utilize the predicted depth map $D$ from \cref{sec:backbone} and the known camera parameters to back-project them into 3D space, generating a sparse 3D anchor point cloud $\mathcal{P}$.

\subsection{3D-Local Attribute Prediction}
\label{sec:prediction}

The second core issue with existing methods is the receptive field mismatch. They employ a large-receptive-field backbone for global context, yet regress 3D Gaussian attributes from a \textbf{single pixel's} feature. We argue this "global-perception, single-point-prediction" design is suboptimal.

\begin{figure}[t]
    \centering
    \includegraphics[width=0.8\linewidth]{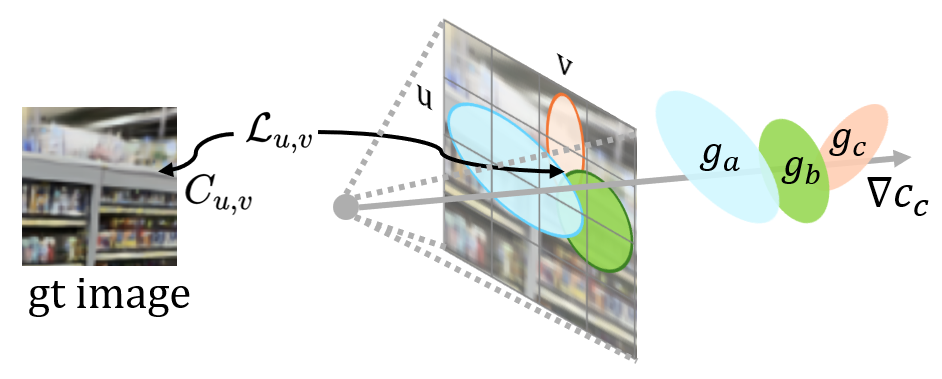}
    \caption{ \textbf{The Locality of Classic 3DGS Optimization.} In this example, three 3D Gaussian primitives are splatted onto the 2D image plane. Primitive $g_c$ covers two pixels: one covered exclusively by $g_c$, and another accumulating contributions from all three primitives. During backpropagation, gradients propagate to $g_c$ through both pixels. Notably, $g_a$ and $g_b$ modulate the gradient flow at the shared pixel by affecting the rendering process. The gradient flow to $g_c$ is detailed in \eqref{eq:gs_description}. }
    \label{fig:grad_flow}
\end{figure}

\paragraph{Design Justification: Locality of 3DGS Optimization.}
We revisit the optimization process of classic 3DGS~\cite{3dgs} as shown in \cref{fig:grad_flow}, a 3D Gaussian primitive is splatted onto the 2D image plane, covering a local region of pixels. The rendering loss (\eg, MSE) is computed on these pixels. Consequently, gradients flow back \textbf{only} from this \textbf{local pixel neighborhood} to the attributes (position, covariance, color) of that single 3D Gaussian. In addition, neighboring 3D Gaussians that have overlap with the 3D Gaussian primitive after 2D projection also affects the rendering process of these pixels, which impacts the gradient flow to the 3D Gaussian primitive. This clearly indicates that 3DGS optimization is an inherently local process. The attributes of a Gaussian is primarily determined by itself and its neighbors in both 3D and 2D space.

\begin{align}
\mathcal{L}_{u,v} &= 
    f_{u,v}(\text{\color[RGB]{70,177,225}{g}}_\text{\color[RGB]{70,177,225}{a}})
    \,\text{\color[RGB]{70,177,225}{c}}_\text{\color[RGB]{70,177,225}{a}}
    + \left(1 - f_{u,v}(\text{\color[RGB]{70,177,225}{g}}_\text{\color[RGB]{70,177,225}{a}})\right)
      f_{u,v}(\text{\color[RGB]{46,139,87}{g}}_\text{\color[RGB]{46,139,87}{b}})
      \,\text{\color[RGB]{46,139,87}{c}}_\text{\color[RGB]{46,139,87}{b}}
      \notag \\
&\quad + \prod_{g \in \{
        \text{\color[RGB]{70,177,225}{g}}_\text{\color[RGB]{70,177,225}{a}},
        \text{\color[RGB]{46,139,87}{g}}_\text{\color[RGB]{46,139,87}{b}}
      \}}
      \left(1 - f_{u,v}(g)\right)
      f_{u,v}(\text{\color[RGB]{242,170,132}{g}}_\text{\color[RGB]{242,170,132}{c}})
      \,\text{\color[RGB]{242,170,132}{c}}_\text{\color[RGB]{242,170,132}{c}}
      - C_{u,v} \notag \\
\nabla c_c &= 
    \mathcal{L}_{u,v} \cdot
    \prod_{g \in \{
        \text{\color[RGB]{70,177,225}{g}}_\text{\color[RGB]{70,177,225}{a}},
        \text{\color[RGB]{46,139,87}{g}}_\text{\color[RGB]{46,139,87}{b}}
     \}}
      (1 - f_{u,v}(g))\,
      f_{u,v}(\text{\color[RGB]{242,170,132}{g}}_\text{\color[RGB]{242,170,132}{c}})  \label{eq:gs_description} \\
&\quad + \mathcal{L}_{u-1,v} \cdot
      f_{u-1,v}(\text{\color[RGB]{242,170,132}{g}}_\text{\color[RGB]{242,170,132}{c}})
      \notag
\end{align}

\paragraph{3D-Local Predictor.}
Based on this finding, we design a lightweight predictor that strictly adheres to this locality principle. For each 3D anchor point $p_i$ in the sparse cloud $\mathcal{P}$, we first query its K-Nearest-Neighbors (KNN) \cite{knn,faiss} in 3D space using efficient FAISS-based search, yielding $K$ neighbors $\{p_{n_1}, ..., p_{n_K}\} \subset \mathcal{P}$. We then extract the corresponding 2D features from the feature map $\mathbf{F}$ for both the central point $p_i$ and its $K$ neighbors. To effectively process the rich information associated with each point, we employ a dual projection strategy. Specifically, for each point (either the center $p_i$ or a neighbor $p_{n_j}$), we denote its geometric features as $\mathbf{g} \in \mathbb{R}^{d_g}$ (comprising xyz coordinates, surface normals, viewing rays) and its high-dimensional image features as $\mathbf{v} \in \mathbb{R}^{d_v}$ (extracted from the backbone). We project them through separate learnable projection functions $\phi_g: \mathbb{R}^{d_g} \to \mathbb{R}^{d_h}$ and $\phi_v: \mathbb{R}^{d_v} \to \mathbb{R}^{d_h}$, then concatenate them to form a unified feature representation:
\begin{equation}
\mathbf{f} = \left[ \phi_g(\mathbf{g}) ; \phi_v(\mathbf{v}) \right] \in \mathbb{R}^{2d_h}
\label{eq:dual_projection}
\end{equation}
where $[\cdot ; \cdot]$ denotes concatenation. This process is applied to both the central point and all its neighbors, yielding an enriched local neighborhood feature set:
\begin{equation}
\mathcal{F}_i = \{\mathbf{f}_i\} \cup \{\mathbf{f}_{n_j}\}_{j=1}^{K}
\label{eq:neighborhood_features}
\end{equation}
which encodes both geometric and appearance cues from the $K$-nearest neighbors.

Given these local neighborhood features, we need to predict the Gaussian attributes for the central point $p_i$. We explored various strategies, including DGCNN-style edge convolutions \cite{dgcnn}, PointNet-style max pooling \cite{qi2017pointnet}, simple MLP, and geo-aware attention \cite{zhao2021point}. Through extensive ablation studies (detailed in \cref{sec:experiment}), we found that geo-aware attention achieves the best performance. Our final predictor employs an attention-based architecture to aggregate information from the $K$ neighbors, allowing the network to weight the contributions of different neighbors adaptively:
\begin{equation}
\tilde{\mathbf{f}}_i = \text{Attention}(\mathbf{f}_i, \mathcal{F}_i)
\label{eq:attention_agg}
\end{equation}
The aggregated feature $\tilde{\mathbf{f}}_i$ is then passed through a lightweight MLP that regresses all necessary Gaussian attributes for the central point $p_i$:
\begin{equation}
\{\alpha, \mathbf{s}, \mathbf{q}, \mathbf{c}\} = \text{MLP}_\theta(\tilde{\mathbf{f}}_i)
\label{eq:attribute_pred}
\end{equation}
where $\alpha \in \mathbb{R}$ is the opacity, $\mathbf{s} \in \mathbb{R}^3$ is the 3D scale, $\mathbf{q} \in \mathbb{R}^4$ is the 3D rotation (as a quaternion), $\mathbf{c} \in \mathbb{R}^{N_{sh} \times 3}$ are the Spherical Harmonics (SH) coefficients, and $\theta$ denotes the learnable parameters of the output MLP. This design offers a dual advantage: 1) it perfectly aligns the predictor's receptive field (a 3D local neighborhood) with the task's inherent nature (local optimization), and 2) it maintains computational efficiency while achieving effective feature aggregation.

\subsection{Training Objective}
\label{sec:training}

Our SparseSplat network is trained end-to-end with RGB supervision only. During training, the predicted set of 3D Gaussian primitives $\mathcal{G}$ is rendered into training views using the standard differentiable Gaussian renderer $\mathcal{R}$~\cite{3dgs}, producing the rendered image $I_{\text{render}}$, and rendered depth $D_{\text{render}}$.

As the depth map $D$ is provided by the pre-trained backbone, our training objective focuses solely on RGB rendering quality. We employ a combination of the Mean Squared Error (MSE) loss $\mathcal{L}_{\text{MSE}}$ and the perceptual LPIPS loss~\cite{lpips} $\mathcal{L}_{\text{LPIPS}}$ to supervise the rendered image $I_{\text{render}}$ against the ground-truth image $I_{\text{gt}}$:
\begin{equation}
\mathcal{L} = (1 - \lambda) \mathcal{L}_{\text{MSE}}(I_{\text{render}}, I_{\text{gt}}) + \lambda \mathcal{L}_{\text{LPIPS}}(I_{\text{render}}, I_{\text{gt}})
\end{equation}
where $\mathcal{L}_{\text{MSE}}$ is the pixel-wise squared error and $\mathcal{L}_{\text{LPIPS}}$ provides a perceptual metric. The hyperparameter $\lambda$ balances these two terms.

\section{Experiments}
\label{sec:experiment}
\subsection{Experimental Setup}
\paragraph{Datasets}
Our main experiments are conducted on the \textbf{DL3DV} \cite{dl3dv} dataset, a widely-used dataset consisting of both real-world indoor and outdoor scenes. We utilize the first 6,000 scenes from DL3DV for training and evaluate on the official test set comprising 140 scenes. We ensure that all scenes in the test set are strictly excluded from the training data. To evaluate generalization capabilities, we also test on the \textbf{Replica} \cite{replica19arxiv} dataset. To the best of our knowledge, this dataset has not been previously employed for training feedforward 3DGS methods. It consists of 28 scenes, all of which are used exclusively for testing.

\paragraph{Evaluation Metrics}
We primarily assess the quality of novel view synthesis using three standard metrics: \textbf{PSNR}, \textbf{SSIM} \cite{wang2004image}, and \textbf{LPIPS} \cite{lpips}. The input views and the rendered novel views are disjoint sets. For efficiency evaluation, we quantify two key aspects: 1) \textbf{Inference Time} and 2) the size of the generated representation, measured by the \textbf{Number of 3D Gaussians} (i.e., Gaussian Map size).

\paragraph{Baselines}
We compare our method against several state-of-the-art open-source baselines. For rendering quality, we focus our comparison on \textbf{MVSplat}~\cite{mvsplat} and \textbf{DepthSplat}~\cite{depthsplat}. These "pixel-aligned" methods represent the current state-of-the-art within the feed-forward 3DGS paradigm. For memory efficiency (Gaussian Map size), we compare against methods including \textbf{GGN}~\cite{GGN}, \textbf{LongLRM}~\cite{longLRM}, and \textbf{AnySplat}~\cite{anysplat}. For DepthSplat and AnySplat, which provide official pre-trained models on DL3DV, we directly evaluate their released weights. For all other methods, we retrain their models on the DL3DV training set, strictly following their prescribed training configurations before evaluation. It is noteworthy that AnySplat does not require ground truth poses; it internally estimates poses using a VGGT backbone~\cite{vggt}. The resulting estimated poses and depth maps are self-consistent, which makes it non-trivial to modify the method to accept ground truth poses to align with our evaluation setup. However, AnySplat is the only open-source baseline capable of adjusting the number of output Gaussians. We therefore include it in our main experiments for a comprehensive efficiency comparison.

\paragraph{Implementation and Training Details}
All models are trained on four NVIDIA A100 (80GB) GPUs. Our final model for the main experiments is trained for approximately 48 hours. We use PyTorch 2.4.1 with CUDA 12.4. The training employs the Adam optimizer combined with a cosine learning rate scheduler. We use a batch size of 2 and a learning rate of $2 \times 10^{-4}$. The entropy window size $N$ is set to 7. For the DL3DV dataset, we use the 270$\times$480 resolution version, resizing all images to 256$\times$448 during both training and evaluation. A minor exception is made for AnySplat, which requires input dimensions to be a multiple of 14; for this baseline, we use a 252$\times$448 resolution.

Crucially, during our training, we \textbf{freeze the weights of the backbone network} that provides initial depth maps and 2D features. This ensures a fair and direct comparison with our primary baseline, DepthSplat, from which the backbone architecture is adopted.

\subsection{Quantitative Results on DL3DV}
We present our main results on the DL3DV test set in Table~\ref{tab:main_results}. Our method, SparseSplat, is evaluated at various operating points with a single model by adjusting the temperature parameter $\tau$ of our adaptive sampling, resulting in different Gaussian counts (150k, 100k, 40k, 10k).

\begin{table}[h!]
\centering
\caption{Quantitative comparison on the DL3DV dataset. $\uparrow$ indicates higher is better, $\downarrow$ indicates lower is better. "GS Cnt" denotes the average number of Gaussian primitives across scenes. }
\label{tab:main_results}
\resizebox{\columnwidth}{!}{%
\begin{tabular}{ll ccc r c}
\toprule
\textbf{Method} & \textbf{Category} & \textbf{PSNR} $\uparrow$ & \textbf{SSIM} $\uparrow$ & \textbf{LPIPS} $\downarrow$ & {\textbf{GS Cnt} $\downarrow$} & \textbf{Time} (s) $\downarrow$ \\
\midrule
MVSplat & pixel-aligned & 22.95 & 0.774 & 0.192 & 688k & 0.260 \\
DepthSplat & pixel-aligned & 24.17 & 0.816 & 0.152 & 688k & 0.128 \\
\midrule
GGN & postprocess & 20.23 & 0.570 & 0.268 & 162k & 0.320 \\
Long-LRM & postprocess & 20.92 & 0.627 & 0.265 & 200k & 0.115 \\
\midrule
AnySplat & Voxelization & 17.45 & 0.471 & 0.320 & 608k & 0.378 \\
 & & 17.34 & 0.463 & 0.330 & 528k & 0.384 \\
 & & 16.14 & 0.417 & 0.390 & 393k & 0.415 \\
 & & 12.22 & 0.309 & 0.519 & 222k & 0.441 \\
 & & 8.91 & 0.239 & 0.618 & 113k & 0.473 \\
\midrule
Ours & Adaptive & 24.20 & 0.817 & 0.168 & 150k & 0.398 \\
 & & 23.95 & 0.786 & 0.189 & 100k & 0.192 \\
 & & 22.65 & 0.737 & 0.251 & 40k & 0.111 \\
 & & 21.29 & 0.665 & 0.321 & 10k & 0.105 \\
\bottomrule
\end{tabular}%
}
\end{table}
\begin{figure*}
    \centering
    \includegraphics[width=1.0\linewidth]{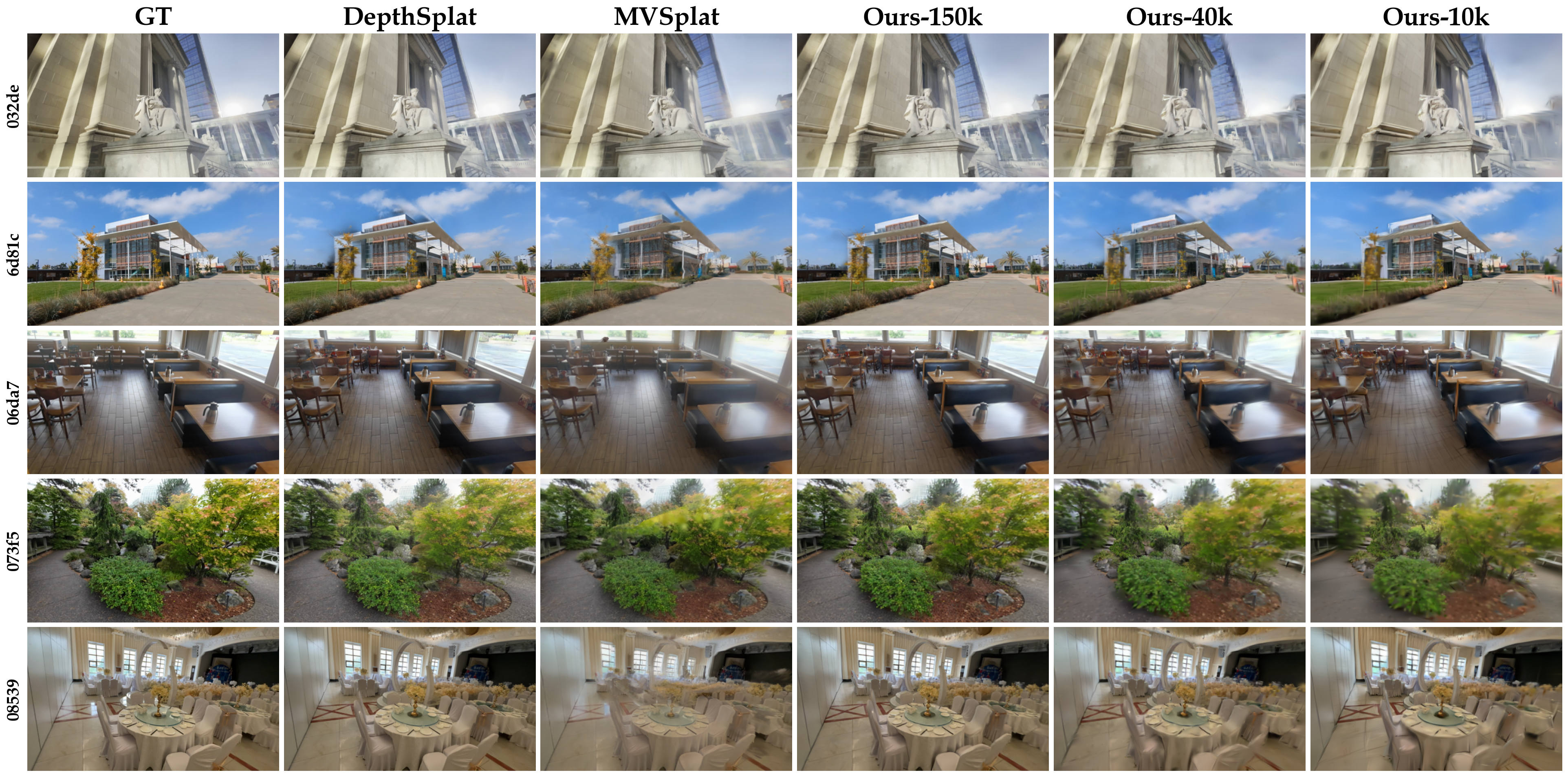}
    \caption{\textbf{Rendering quality comparisons on DL3DV.} Our model matches the SOTA rendering quality of DepthSplat with only 150k Gaussians (vs. 688k). Under sparse settings (40k and 10k), our method maintains structural integrity and shows minor progressive blurring. }
    \label{fig:comparison}
\end{figure*}

\paragraph{Quantitative Results and Applicability}
Our main results on the DL3DV test set (Table~\ref{tab:main_results}) demonstrate SparseSplat's ability to produce highly efficient and flexible scene representations. 

\textbf{High-Fidelity, Compact Models.} Our 150k model achieves state-of-the-art rendering quality, scoring a \textbf{24.20} PSNR which is on par with the DepthSplat baseline (24.17). Crucially, it achieves this using only \textbf{150k} Gaussians compared to DepthSplat's 688k. This significant reduction in memory footprint involves a time trade-off; the 150k model's reconstruction time (398ms) is slower than the baseline (128ms), mainly due to the KNN process. However, it is still suited for applications like \textbf{AR/VR}, where a higher one-time reconstruction cost is acceptable to gain a highly compact model (small memory footprint) ideal for efficient on-device storage and real-time rendering.

\textbf{High-Speed, Sparse Models.} SparseSplat's core strength is its flexibility. For real-time applications like \textbf{3DGS-SLAM}, which require extreme sparsity for sequential updates, our 10k and 40k models are an ideal choice. They are both faster than the DepthSplat baseline (105ms/111ms vs 128ms) and maintain robust quality (21.29 and 22.65 PSNR, respectively). This ability to degrade gracefully in sparse settings contrasts with other methods like AnySplat, which suffers severe performance drops.

\subsection{Cross-Dataset Generalization on Replica}
To test generalization, we evaluate our model trained on DL3DV directly on the unseen Replica dataset. The results are compared against DepthSplat and MVSplat in Table~\ref{tab:replica}.

\begin{table}[h!]
\centering
\caption{Cross-dataset generalization results on the Replica dataset.}
\label{tab:replica}
\resizebox{0.46\textwidth}{!}{ 
\begin{tabular}{l >{\centering\arraybackslash}p{1.7cm} >{\centering\arraybackslash}p{1.7cm} >{\centering\arraybackslash}p{2.1cm}}
\toprule
\textbf{Method} & \textbf{PSNR} $\uparrow$ & \textbf{SSIM} $\uparrow$ & \textbf{LPIPS} $\downarrow$ \\
\midrule
MVSplat & 19.13 & 0.628 & 0.423 \\
DepthSplat & 26.47 & 0.836 & \textbf{0.175} \\
Ours-150k & \textbf{26.64} & \textbf{0.846} & 0.180 \\
\bottomrule
\end{tabular}
}
\end{table}

As shown in Table~\ref{tab:replica}, our SparseSplat model demonstrates strong generalization capabilities. It achieves a PSNR of 26.639, surpassing DepthSplat (26.471). While DepthSplat maintains a slight edge in LPIPS, our superior PSNR score indicates that our model can effectively reconstruct novel scenes from entirely different domains without re-training. MVSplat, in contrast, shows poor generalization.

\subsection{Qualitative Analysis}We provide qualitative comparisons in Figure \ref{fig:comparison} on five scenes randomly sampled from the DL3DV test set. Our method demonstrates superior or comparable rendering quality to existing state-of-the-art feed-forward methods while operating at a fraction of the primitive count.\paragraph{High-Fidelity Sparse Reconstruction.} In typical scenes such as 032de (statue), 073f5 (garden), and 08539 (banquet hall), our model (Ours-150k) achieves rendering quality that is comparable or even superior to DepthSplat. This is achieved while using only approximately 22\% of the Gaussian primitives (150k vs. 688k for DepthSplat), validating the effectiveness of our adaptive primitive sampling (\cref{sec:sampling}). Furthermore, our model degrades gracefully. Even when the primitive count is drastically reduced to 40k or 10k, SparseSplat maintains high fidelity and structural integrity, exhibiting only minor, progressive blurring.

\paragraph{Robustness to Geometric Errors.} The 6d81c (building) scene shows an advantage of our design. This scene suffers from depth estimation errors from the backbone at the sky-house boundary, which cause rendering degradation in the DepthSplat baseline. Despite utilizing the same frozen depth backbone, SparseSplat is more robust to these geometric inaccuracies. We attribute this to our 3D-Local Attribute Predictor (\cref{sec:prediction}). Unlike "pixel-aligned" methods that predict attributes from a single pixel's feature, our predictor processes a 3D neighborhood. This allows it to more robustly identify local geometric inconsistencies. Also, the sparse nature allows our model to mitigate the impact of depth errors by predicting smaller, more localized Gaussian primitives in the erroneous regions, preventing the artifact from propagating.


\subsection{Ablation Studies}
We conduct a series of ablation studies to validate our core design choices: the adaptive sampling strategy and the 3D-local attribute predictor. For computational efficiency, all models in this section are trained for 30k steps on four A100 GPUs and evaluated using a 40k Gaussian count (GS Cnt) setting.

\subsubsection{Ablation on Sampling Strategy}
We compare our proposed entropy-based adaptive sampling against two alternatives: (1) \textbf{Random}: uniform random sampling of pixels, and (2) \textbf{Laplacian}: sampling based on image gradients (Laplacian filter), which is a common proxy for edges \cite{theoryforedge}. The results in Table~\ref{tab:ablation_sampling} show that our entropy-based method (22.36 PSNR) significantly outperforms both random (21.50 PSNR) and Laplacian-based (21.95 PSNR) sampling. This validates our hypothesis that Shannon entropy is a more effective measure of "information richness" for guiding sparse Gaussian allocation than simple edge detection or random chance, directly addressing the "distribution mismatch" problem.
\vspace{-6pt}
\begin{table}[h!]
\centering
\caption{Ablation on the primitive sampling strategy.}
\label{tab:ablation_sampling}
\begin{tabular}{l ccc}
\toprule
\textbf{Sampling Strategy} & \textbf{PSNR} $\uparrow$ & \textbf{SSIM} $\uparrow$ & \textbf{LPIPS} $\downarrow$ \\
\midrule
\textbf{Entropy (Ours)} & \textbf{22.36} & \textbf{0.718} & \textbf{0.262} \\
Random & 21.50 & 0.696 & 0.288 \\
Laplacian & 21.95 & 0.705 & 0.267 \\
\bottomrule
\end{tabular}
\end{table}

\subsubsection{Ablation on 3D-Local Predictor Design}
We validate two key components of our 3D-Local Attribute Predictor: the K-Nearest-Neighbors (KNN) query and the feature aggregation head.

\paragraph{Effect of K in KNN} We vary the number of neighbors (K) used by the predictor. As shown in Table~\ref{tab:ablation_k}, performance degrades significantly when K=0 (PSNR 21.53), which reduces our model to a simple MLP operating on a single point. This confirms that aggregating local neighborhood information is critical. Performance increases steadily with K, with diminishing returns after K=20. This shows that local neighborhood information is beneficial for the prediction of 3DGS attributes, which supports the "receptive field mismatch" hypothesis. We choose K=20 (22.36 PSNR) as our default setting, balancing performance and computational cost.
\begin{table}[h!]
\centering
\caption{Effect of K in KNN.}
\label{tab:ablation_k}
\resizebox{0.46\textwidth}{!}{ 
\begin{tabular}{l >{\centering\arraybackslash}p{1.8cm} >{\centering\arraybackslash}p{1.8cm} >{\centering\arraybackslash}p{1.9cm}}
\toprule
\textbf{K} & \textbf{PSNR} $\uparrow$ & \textbf{SSIM} $\uparrow$ & \textbf{LPIPS} $\downarrow$ \\
\midrule
0 & 21.53 & 0.683 & 0.291 \\
5 & 22.23 & 0.713 & 0.267 \\
10 & 22.34 & 0.717 & 0.264 \\
20 & 22.36 & \textbf{0.718} & 0.262 \\
30 & \textbf{22.38} & \textbf{0.718} & \textbf{0.261} \\
\bottomrule
\end{tabular}
}
\end{table}

\paragraph{Effect of Prediction Head} As mentioned in \ref{sec:method}, we evaluated several standard point-based network architectures for the local feature aggregation task. The results are presented in Table~\ref{tab:ablation_head}. We found that the geo-aware attention mechanism (22.36 PSNR) substantially outperforms other methods. Furthermore, we observed that a PointNet-style max-pooling head consistently failed during training.  This ablation study validates our selection of the geo-aware attention mechanism to aggregate local 3D neighborhood features.

\begin{table}[h!]
\centering
\caption{Effect of Prediction Head.}
\label{tab:ablation_head}
\begin{tabular}{l ccc}
\toprule
\textbf{Head} & \textbf{PSNR} $\uparrow$ & \textbf{SSIM} $\uparrow$ & \textbf{LPIPS} $\downarrow$ \\
\midrule
Graph-Conv & 20.78 & 0.664 & 0.329 \\
MLP & 21.47 & 0.683 & 0.299 \\
Max Pooling & Fail & Fail & Fail \\
\textbf{Geo-aware Attn} & \textbf{22.36} & \textbf{0.718} & \textbf{0.263} \\
\bottomrule
\end{tabular}
\end{table}
\vspace{-4pt}
\section{Conclusion}
\label{sec:conclusion}
We present SparseSplat, a feed-forward 3DGS framework that addresses the redundancy of "pixel-aligned" methods. Our approach resolves the distribution and receptive field mismatches inherent in prior work using two key components: (1) entropy-based adaptive sampling to dynamically allocate Gaussians and (2) a 3D-local attribute predictor to align with 3DGS's local optimization nature. Experimental results show that SparseSplat achieves state-of-the-art rendering quality while using significantly fewer Gaussians. SparseSplat significantly improves efficiency and makes feed-forward 3DGS practical for downstream applications like SLAM \cite{3dgsslam_survey}, AR/VR \cite{ARVR_survey}, and robotics \cite{robot_survey}.

\paragraph{Limitations} While our 3D-Local Attribute Predictor demonstrates robustness to minor geometric errors, 3D KNN might not be the optimal way to gather useful context under severe depth estimation errors. For example, if a point from view A is incorrectly projected into the foreground of view B and completely occludes the underlying scene, 3D KNN will fail to include points from the occluded regions of view B as context for that point. However, those occluded points are important cues indicating that the point is positioned incorrectly. We believe that a context aggregation mechanism based on 2D co-visibility would be more effective. Another limitation of the KNN-based context aggregation is its computational inefficiency. In future work, we will explore the possibility of a more efficient context-aggregation strategy based on 2D co-visibility.
\newpage
\section*{Acknowledgments}
This work was supported in part by the National Natural Science Foundation of China (NSFC) under Grant 62403142, and in part by the Science and Technology Commission of Shanghai Municipality under Grant 24511103100. We would also like to thank Tiancheng Wu from Carnegie Mellon University for his valuable suggestions on point cloud processing networks.
{
    \small
    \bibliographystyle{ieeenat_fullname}
    \bibliography{main}
}

\clearpage
\setcounter{page}{1}
\twocolumn[{
    \renewcommand\twocolumn[1][]{#1} 
    \maketitlesupplementary           
    
    \begin{center}
        \centering
        \vspace{-15pt} 
        \includegraphics[width=1.0\linewidth]{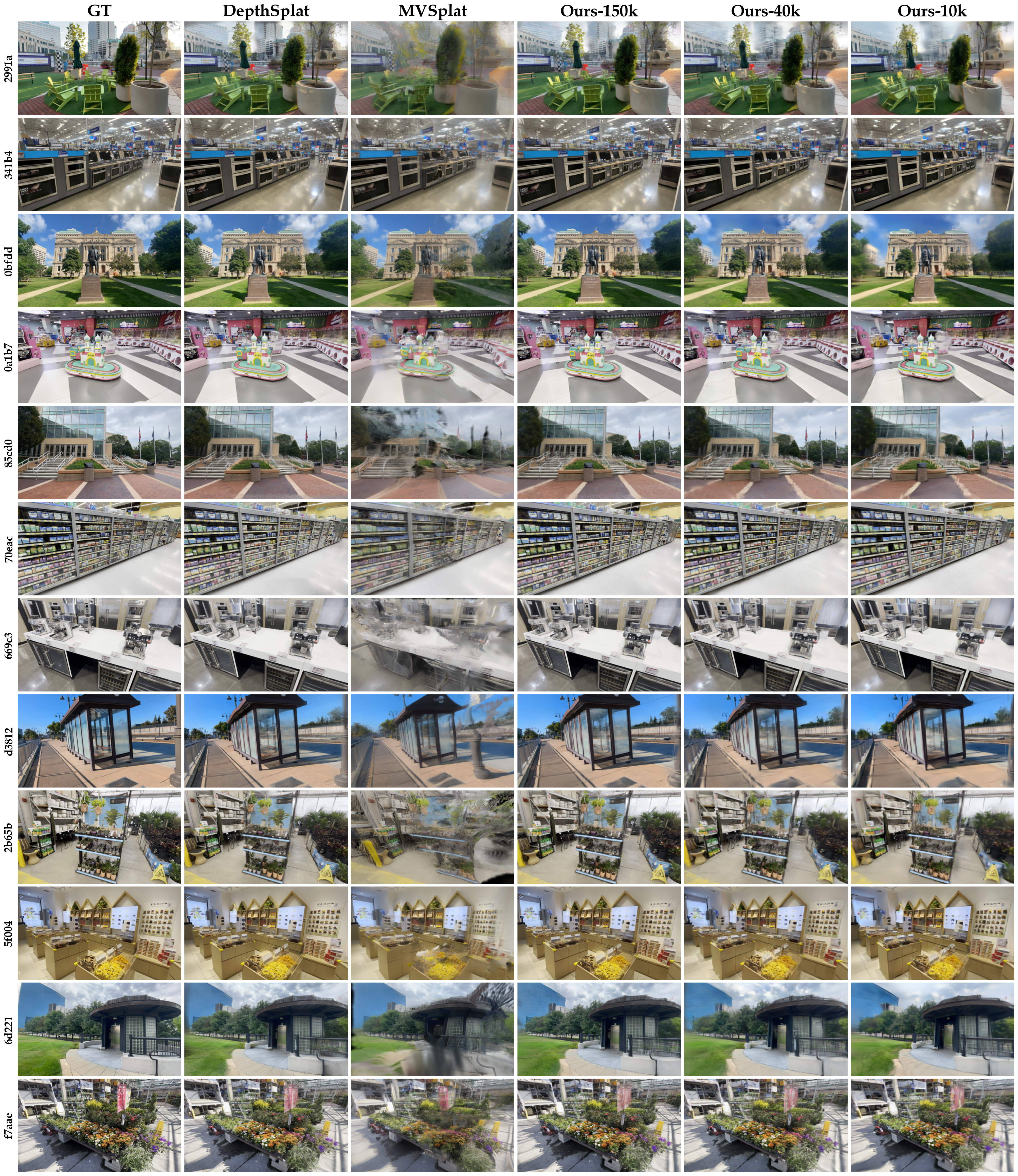}
        
        \captionof{figure}{\textbf{Additional qualitative comparisons.} }
        \label{fig:more_comp}
    \end{center}
}]

\noindent This supplementary material provides additional quantitative and qualitative results, along with an assessment of SparseSplat's applicability to downstream scenarios. We also present architectural details of the prediction heads and confirm that our sparse-by-design representation delivers strong accuracy–efficiency trade-offs across varying Gaussian budgets.

\section{Applicability to Downstream Tasks}
To demonstrate the practical value of SparseSplat, we evaluate how it integrates into various downstream tasks. We categorize these tasks based on two different forms of ``real-time'' requirements: \textbf{Reconstruction Real-time-ness}, which underpins online mapping and robotic perception, and \textbf{Rendering Real-time-ness}, which is indispensable for immersive AR/VR experiences and large-scale simulation. Owing to its sparse-by-design architecture, SparseSplat naturally accommodates both types of real-time constraints, whereas pixel-aligned feed-forward methods struggle to satisfy them simultaneously due to their rigid, dense Gaussian allocation.

\begin{figure}[htbp]
    \centering
    \includegraphics[width=\linewidth]{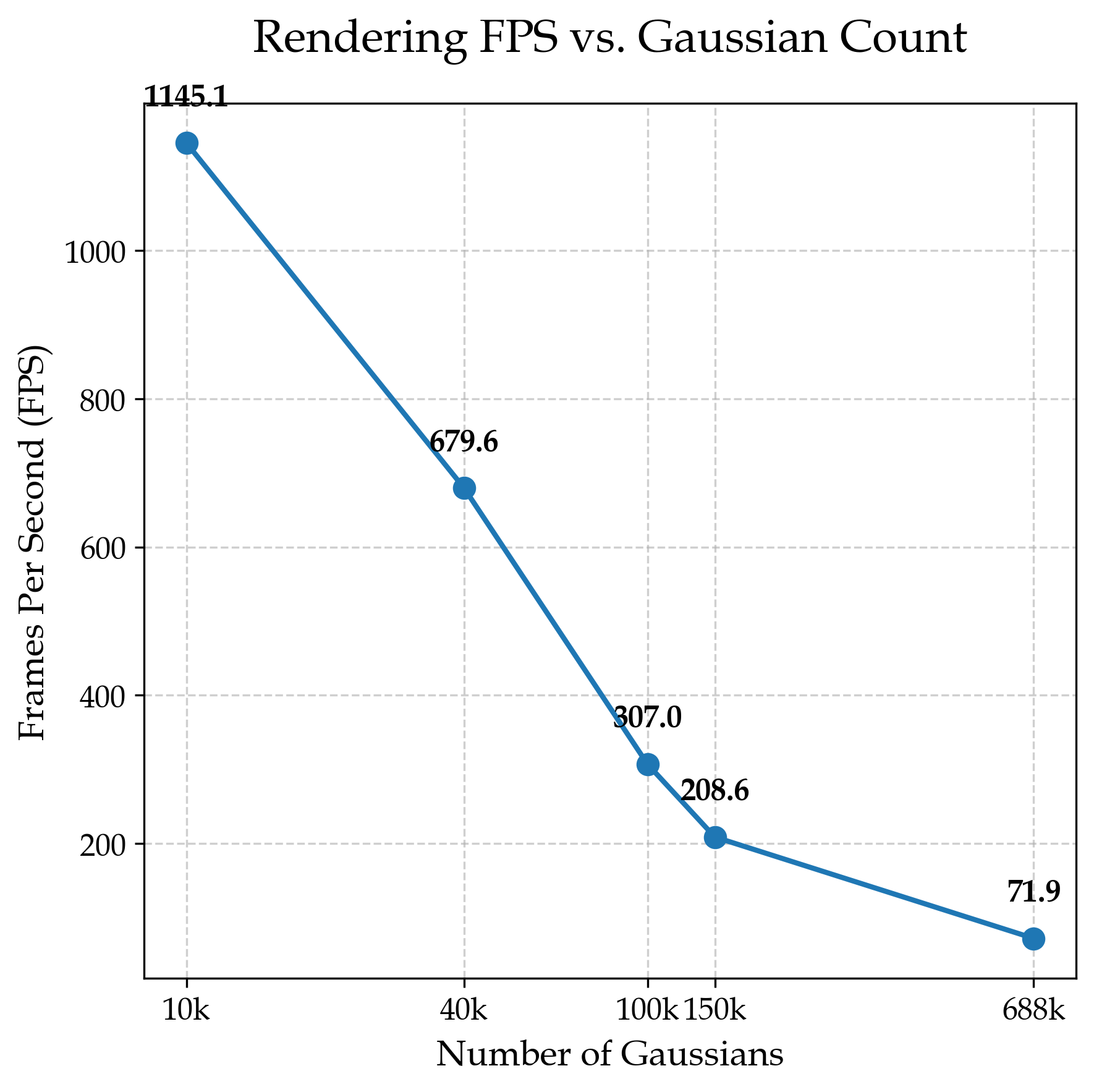}
    \caption{\textbf{Rendering Efficiency vs. Gaussian Count. X-axis log-scaled.} We evaluate the rendering frame rate (FPS) across scenes with varying degrees of sparsity. While pixel-aligned baselines (typified by $\sim$688k Gaussians) operate at 71.9 FPS, our sparse-by-design approach significantly accelerates rendering. Our 150k model achieves $\sim$3$\times$ speedup (208.6 FPS), and extremely sparse settings (10k--40k) unlock rates suitable for high-frequency robotics control loops ($>$600 FPS).}
    \label{fig:fps_vs_gs}
\end{figure}
\subsection{Reconstruction Real-time-ness: 3DGS-SLAM}
\paragraph{Context.}
Simultaneous Localization and Mapping (SLAM) requires the online construction of 3D scenes from continuous video streams. Leveraging 3D Gaussian Splatting, recent methods like GS-SLAM~\cite{MonoGS} and SplaTAM~\cite{splatam} enable faster updates and better reconstruction than prior baselines. A crucial objective for these systems is controlling the primitive count: as shown in \cref{fig:fps_vs_gs}, rendering latency is strictly tied to the Gaussian budget, where lower counts ensure real-time performance.

\paragraph{Integration and Sparsity.}
Feed-forward methods can be seamlessly integrated into these pipelines to replace computationally expensive per-frame optimization. Specifically, given a window of $N$ incoming frames and their estimated poses (from the tracking thread), SparseSplat directly predicts a set of 3D Gaussians.

However, a critical bottleneck for existing feed-forward methods (e.g., DepthSplat~\cite{depthsplat}, MVSplat~\cite{mvsplat}) is their ``pixel-aligned'' nature. They rigidly generate a Gaussian for every pixel, adding 688k primitives per inference. In one SLAM sequence with thousands of frames, this rapid accumulation leads to memory explosion and tracking lag, effectively rendering them unusable for long-sequence operation.
In contrast, SparseSplat employs a ``sparse-by-design'' representation. By adaptively allocating primitives only where necessary (e.g., 10k--40k per view), our method significantly reduces memory consumption and update overhead, enabling robust operation on resource-constrained platforms where pixel-aligned baselines fail.

\begin{table}[h]
\centering
\caption{Quantitative comparison of Gaussian primitive counts across different methods and datasets. Data sourced from CaRtGS.} 
\label{tab:primitive_counts}
\resizebox{\linewidth}{!}{%
\begin{tabular}{lllc}
\toprule
\textbf{Dataset} & \textbf{Method} & \textbf{Sensor} & \textbf{Gaussian Primitives (M)} \\
\midrule
\multirow{3}{*}{\textbf{Replica} (office0)} 
 & GS-ICP-SLAM & RGB-D & 1.57 M \\
 & Photo-SLAM & RGB-D & 0.08 M \\
 & Photo-SLAM & Monocular & 0.08 M \\
\midrule
\multirow{5}{*}{\textbf{TUM-RGBD} (fr1/desk)} 
 & GS-ICP-SLAM & RGB-D & 1.56 M \\
 & SplaTAM & RGB-D & 0.96 M \\
 & Gaussian-SLAM & RGB-D & 0.70 M \\
 & MonoGS & Monocular & 0.04 M \\
 & Photo-SLAM & Monocular & 0.02 M \\
\midrule
\multirow{5}{*}{\textbf{TUM-RGBD} (fr3/office)} 
 & GS-ICP-SLAM & RGB-D & 1.09 M \\
 & Gaussian-SLAM & RGB-D & 1.04 M \\
 & SplaTAM & RGB-D & 0.79 M \\
 & MonoGS & Monocular & 0.03 M \\
 & Photo-SLAM & Monocular & 0.01 M \\
\bottomrule
\end{tabular}%
}
\end{table}
\paragraph{Scalability Analysis.}
A potential concern regarding our method is the use of K-Nearest Neighbors (KNN) for local attribute prediction: does the inference time increase unacceptably as the global map grows?
It is important to clarify that while the \textit{global} map accumulates millions of points over time, our inference is local. We only perform KNN queries for the sparse anchor points generated from the current input batch of $N$ frames. Therefore, the computational cost per update step remains constant and low, preserving the real-time nature of the system regardless of the total map size.

\subsection{Rendering Real-time-ness: AR/VR and Simulation}

The second category of applications prioritizes the quality and speed of the \textit{final rendering} rather than the speed of the reconstruction process itself.

\paragraph{AR/VR on Edge Devices.}
In Virtual and Augmented Reality, devices (e.g., headsets, mobile phones) operate under strict VRAM and thermal constraints. A scene represented by 688k Gaussians (e.g., via DepthSplat) imposes a significantly heavier rendering load than one represented by 150k Gaussians (Ours). By offering a 4.5$\times$ reduction in primitive count without sacrificing quality, SparseSplat ensures higher rendering frame rates (FPS) and lower memory footprint, making it ideal for on-device deployment.

\paragraph{High-Throughput Simulation for Robotics.}
In robotics, agents are often trained in simulated environments (e.g., for visual navigation). These training loops require rendering millions of observations. Consequently, the rendering speed of the environment becomes the primary bottleneck for training throughput.
Our method generates highly compact digital assets that render significantly faster than those produced by pixel-aligned methods. This efficiency directly translates to faster simulation speeds, allowing for more efficient large-scale training of visual reinforcement learning algorithms.
\section{Runtime Breakdown}
We present a detailed runtime breakdown of individual components across varying Gaussian counts in ~\cref{tab:time_breakdown}. The latency of backbone inference and entropy-based sampling remains constant regardless of the sparsity level. In contrast, the computational costs of the KNN query and the Attention prediction head scale with the number of generated Gaussians.
\begin{table}[t]
    \centering
    \caption{\textbf{Inference Time Breakdown (ms).} We detail the computational cost of each component in SparseSplat across different Gaussian counts. Note that the total time reported here is slightly higher than in the main text due to the synchronization overhead introduced by time profiling.}
    \label{tab:time_breakdown}
    \resizebox{\linewidth}{!}{
        \begin{tabular}{l|cccc|r}
            \toprule
            \textbf{Model} & \textbf{Backbone} & \textbf{Sampling} & \textbf{KNN} & \textbf{Attention} & \textbf{Total} \\
            \midrule
            Ours-10k  & 89.68 & 1.66 & 3.21   & 5.95   & \textbf{100.50} \\
            Ours-40k  & 89.68 & 1.82 & 14.87  & 15.04  & \textbf{121.41} \\
            Ours-100k & 89.68 & 2.16 & 52.02  & 65.18  & \textbf{209.04} \\
            Ours-150k & 89.68 & 2.08 & 219.64 & 103.31 & \textbf{414.71} \\
            \bottomrule
        \end{tabular}
    }
\end{table}
\section{Structure of Different Heads}
\label{sec:appendix_heads}

As described in \cref{sec:prediction}, our 3D-Local Attribute Prediction framework employs a lightweight predictor to regress Gaussian attributes based on K-nearest neighbors in 3D space. We explored four different prediction head architectures, all sharing the same dual projection strategy for processing geometric and image features but differing in how they aggregate information from the local neighborhood. This section provides structural details of these variants.

\subsection{MLP Head}

For each anchor point, we gather features from its $K$ nearest neighbors and compute their relative positions $\Delta \mathbf{p}_j = \mathbf{p}_j - \mathbf{p}_i$. The aggregated neighborhood features (relative geometry and appearance) are concatenated with the center point's features and fed into a 4-layer MLP to directly regress the Gaussian parameters.

\subsection{DGCNN Head}

The DGCNN head adopts the EdgeConv architecture from \cite{wang2019dynamic}, which dynamically constructs local graphs to capture fine-grained geometric structures. For each anchor point, EdgeConv models pairwise relationships with its neighbors by computing edge features that encode both local geometric context and absolute point features. Specifically, for a center point with feature $\mathbf{f}_i$ and its neighbor $\mathbf{f}_j$, the edge feature is computed as:
\begin{equation}
\mathbf{h}_{ij} = \text{MLP}(\mathbf{f}_i \oplus (\mathbf{f}_j - \mathbf{f}_i))
\end{equation}
where $\oplus$ denotes concatenation, and $\mathbf{f}_j - \mathbf{f}_i$ captures the relative geometric offset. This formulation allows the network to learn both global point characteristics (through $\mathbf{f}_i$) and local geometric variations (through the difference term).

The architecture consists of multiple stacked EdgeConv layers with progressively increasing channel dimensions. After each EdgeConv layer, a symmetric max-pooling operation aggregates edge features across neighbors, ensuring permutation invariance. The features from all EdgeConv layers are then concatenated to form a multi-scale representation that captures geometric patterns at different levels of abstraction. A final prediction head regresses the Gaussian attributes from this rich feature descriptor. 

\subsection{PointNet Head}

The PointNet head is based on the classical PointNet architecture \cite{qi2017pointnet}. It treats the center point and its $K$ neighbors as an unordered set of $K+1$ points. A shared MLP with architecture [128 $\to$ 256 $\to$ 512 $\to$ 1024] processes each point independently, and a max-pooling operation aggregates the point-wise features into a global descriptor. A subsequent prediction MLP [1024 $\to$ 512 $\to$ 256 $\to$ 38] regresses the Gaussian parameters from this descriptor. The shared MLP and symmetric pooling ensure permutation invariance.

\subsection{Geo-Aware Attention Head}

The Geo-Aware Attention head employs a Point Transformer-style vector attention mechanism \cite{zhao2021point}, which we find most effective for this task. Unlike simple pooling-based aggregation, it adaptively weights neighbor contributions based on both feature similarity and geometric relationships.

\paragraph{Position-Aware Vector Attention.}
The core mechanism computes attention weights that incorporate relative positions. For each center point $\mathbf{f}_i$ and its neighbors $\{\mathbf{f}_j\}_{j=1}^K$, we compute queries $\mathbf{Q}_i$, keys $\mathbf{K}_j$, and values $\mathbf{V}_j$ via learned projections. The relative position $\Delta \mathbf{p}_j = \mathbf{p}_j - \mathbf{p}_i$ is encoded through an MLP to obtain position embeddings $\mathbf{\delta}_j$. The attention weights are then computed as:
\begin{equation}
\alpha_{ij} = \text{softmax}_j\left(\frac{(\mathbf{Q}_i - \mathbf{K}_j + \mathbf{\delta}_j) \cdot \mathbf{Q}_i}{\sqrt{d}}\right)
\end{equation}
where the position encoding $\mathbf{\delta}_j$ modulates the attention computation. The output feature is obtained by:
\begin{equation}
\mathbf{f}'_i = \sum_{j=1}^K \alpha_{ij} (\mathbf{V}_j + \mathbf{\delta}_j)
\end{equation}
This design allows the network to learn geometry-aware feature aggregation, where the position encoding influences both attention weights and value features.

Among all examined variants, the Geo-aware attention head provides the best balance between accuracy and efficiency, underscoring the importance of geometry-conditioned local aggregation in SparseSplat.

\paragraph{Architecture.}
We employ three stacked attention layers with hidden dimension 128 and 4 attention heads per layer. Each layer includes a feed-forward network (FFN) with expansion ratio 4, residual connections, and layer normalization. The dual projection strategy separately processes geometric features and image features before concatenation, maintaining modularity between geometry and appearance processing.

%

\end{document}